\documentclass[conference]{IEEEtran}
\usepackage{times}
\usepackage{float}
\usepackage[numbers]{natbib}
\usepackage{amsmath}
\usepackage{mathtools}
\usepackage{booktabs}
\usepackage{multirow}
\usepackage{amssymb}
\usepackage{graphicx}
\usepackage{svg}
\usepackage{pgffor}
\usepackage{booktabs}
\usepackage[export]{adjustbox}
\usepackage{array}    % for *{N}{c} column repetition
\usepackage{caption}
\usepackage[bookmarks=true]{hyperref}

\begin{document}

\title{$\pi$, But Make It Fly: Physics-Guided Transfer of VLA Models to Aerial Manipulation}

% You will get a Paper-ID when submitting a pdf file to the conference system
% \author{Author Names Omitted for Anonymous Review. Paper-ID 1077}

\author{
\IEEEauthorblockN{%
\centering
\begin{tabular}{ccccc}
Johnathan Tucker\textsuperscript{1} &
Denis Liu\textsuperscript{1} &
Aiden Swann\textsuperscript{1} &
Allen Ren\textsuperscript{2} &
Javier Yu\textsuperscript{1} \\
Jiankai Sun\textsuperscript{1} &
Brandon Kim\textsuperscript{1} &
Lachlain McGranahan\textsuperscript{1} &
Quan Vuong\textsuperscript{2} &
Mac Schwager\textsuperscript{1}
\end{tabular}
}
\IEEEauthorblockA{%
\textsuperscript{1}Stanford University\\
\textsuperscript{2}Physical Intelligence\\
\texttt{\url{https://airvla.github.io}}
}
}

\makeatletter
    \let\@oldmaketitle\@maketitle % Store \@maketitle
    \renewcommand{\@maketitle}{
        \@oldmaketitle% % Update \@maketile to insert...
        % \bigskip
        \vspace{1.5ex}

            \includegraphics[width=\linewidth]{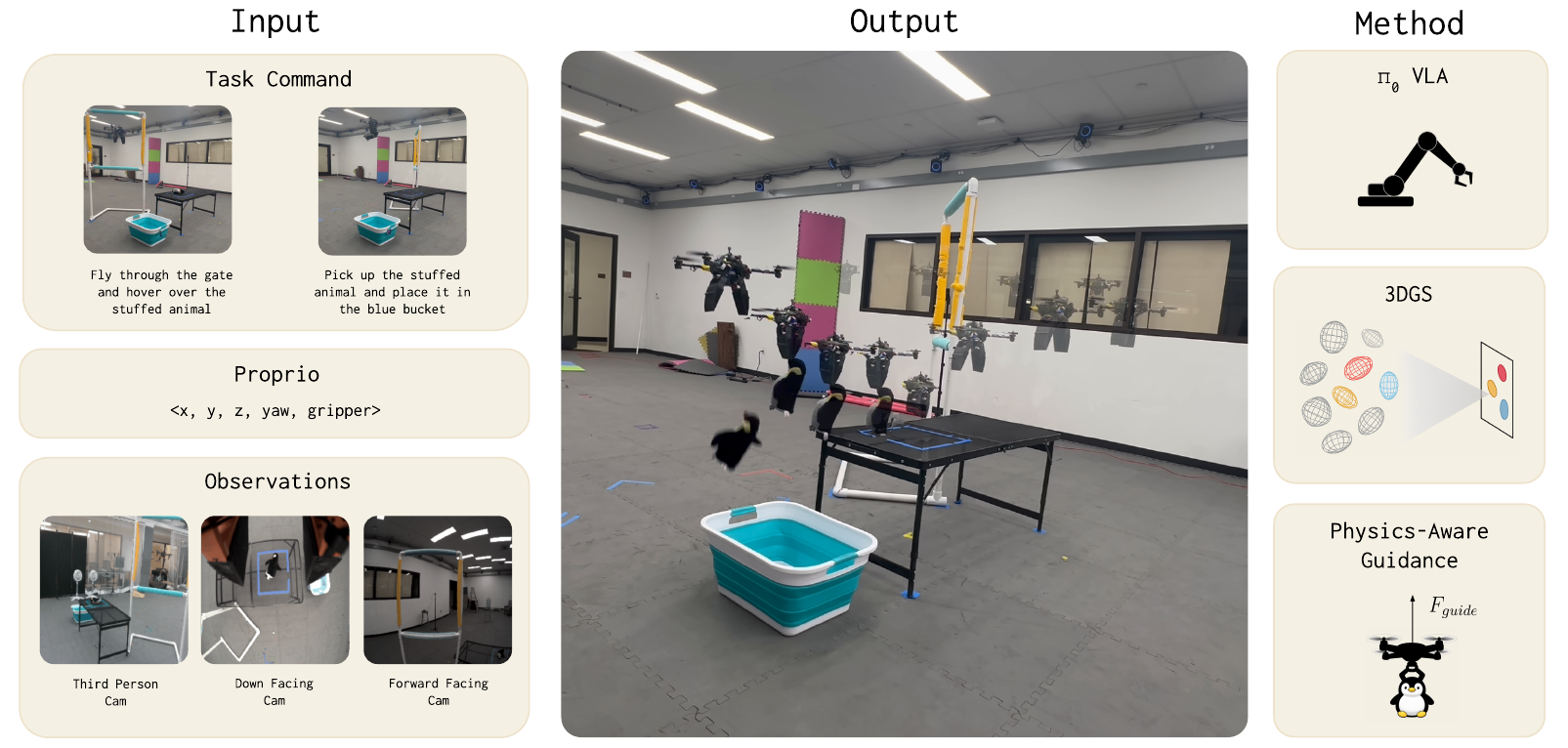}
            % \includesvg[width=\linewidth]{figures/banner/splashv3_dark.svg}
            \captionof{figure}{\textbf{Overview of AirVLA:} Our method fine-tunes the \(\pi_0\) vision-language-action model on a combination of teleoperated and 3D Gaussian Splatting synthetic data. \textbf{(Left)} The policy processes multimodal inputs including natural language commands, proprioception, and multi-view camera observations. \textbf{(Right)} To ensure robust flight during manipulation, we introduce a payload-aware guidance signal, \(F_{guide}\), combined with real-time chunking. \textbf{(Center)} This enables AirVLA to execute novel, zero-shot compositional tasks, such as navigating through gates and manipulating objects.}\label{fig:dronevla}

    }
\makeatother
    
% \maketitle
% \thispagestyle{empty}
% \pagestyle{empty}

\maketitle
\addtocounter{figure}{-1}
\begin{abstract}
Vision-Language-Action (VLA) models such as $\pi_0$ have demonstrated remarkable generalization across diverse fixed-base manipulators. However, transferring these foundation models to aerial platforms remains an open challenge due to the fundamental mismatch between the quasi-static dynamics of fixed-base arms and the underactuated, highly dynamic nature of flight. In this work, we introduce AirVLA, a system that investigates the transferability of manipulation-pretrained VLAs to aerial pick-and-place tasks. We find that while visual representations transfer effectively, the specific control dynamics required for flight do not. To bridge this "dynamics gap" without retraining the foundation model, we introduce a Payload-Aware Guidance mechanism that injects payload constraints directly into the policy’s flow-matching sampling process. To overcome data scarcity, we further utilize a Gaussian Splatting pipeline to synthesize navigation training data. We evaluate our method through a cumulative 460 real-world experiments which demonstrate that this synthetic data is a key enabler of performance, unlocking 100\% success in navigation tasks where directly fine-tuning on teleoperation data alone attains 81\% success. Our inference-time intervention, Payload-Aware Guidance, increases real-world pick-and-place task success from 23\% to 50\%. Finally, we evaluate the model on a long-horizon compositional task, achieving a 62\% overall success rate. These results suggest that pre-trained manipulation VLAs, with appropriate data augmentation and physics-informed guidance, can transfer to aerial manipulation and navigation, as well as the composition of these tasks.
\end{abstract}

\IEEEpeerreviewmaketitle

\section{Introduction}
% What is the problem?
Vision-Language-Action (VLA) foundation models have demonstrated remarkable cross-embodiment generalization, enabling robots with diverse morphologies to complete manipulation tasks from natural language instructions. Models such as RT-X~\cite{open_x_embodiment_rt_x_2023}, Octo~\cite{octo_2023}, and $\pi_0$~\cite{pi-zero_2024} leverage large-scale multimodal pretraining across dozens of robot embodiments, learning representations that transfer across manipulators with different kinematics, workspaces, and end-effector designs. However, all such demonstrations share a fundamental constraint: they operate from stable, fixed or mobile bases in quasi-static regimes. This raises a central question: \emph{can VLA policies transfer to aerial manipulators}, robots that couple perception, language understanding, and contact-rich interaction with underactuated 6-DoF flight control? Such transfer would represent an extreme test of cross-embodiment learning, as aerial platforms differ fundamentally from ground robots in dynamics, control authority, and sensing characteristics.

% Why is it interesting and important?
Answering this question has both scientific and practical significance. Scientifically, aerial manipulation provides a stress test for the representations learned by VLA foundation models: if policies pretrained entirely on fixed-base manipulators can partially transfer to flying robots through fine-tuning, it suggests these models capture manipulation primitives that transcend specific embodiment constraints. Practically, language-conditioned aerial manipulation would enable applications previously beyond reach: delivering medical supplies in inaccessible terrain, clearing debris in collapsed structures, or manipulating infrastructure at height. If VLA policies could be adapted for such platforms, non-expert operators could issue high-level language commands while the policy handles low-level perception and control.

% Why is it hard? (E.g., why do naive approaches fail?)
Several fundamental challenges distinguish aerial manipulation from the ground-robot scenarios that dominate VLA training data. First, quadrotors are underactuated: thrust and attitude are tightly coupled, and small errors in predicted actions can induce large deviations in pose. This failure mode rarely occurs in fixed-base manipulation. Second, onboard cameras experience large ego-motion, rapid viewpoint changes, motion blur, and scale variation that differ significantly from tabletop scenarios in most VLA datasets~\cite{khazatsky2024droid,open_x_embodiment_rt_x_2023}. Finally, contact with objects induces payload changes~\cite{bodie2019omnidirectional,he2023image} and aerodynamic disturbances~\cite{sanchez2017characterization,o2022neural} that violate quasi-static assumptions implicit in most manipulation policies.

% Why hasn't it been solved before? (Or, what's wrong with previous proposed solutions? How does mine differ?)
While recent work has begun exploring language-conditioned aerial systems and cross-embodiment transfer to drones, no prior work has systematically investigated whether manipulation-pretrained VLA foundation models can transfer to aerial platforms. Concurrent efforts take fundamentally different approaches: UMI-on-Air~\cite{gupta2025umi} trains embodiment-agnostic diffusion policies from scratch on handheld demonstrations and introduces inference-time guidance to compensate for aerial dynamics, but does not leverage or evaluate VLA foundation models. SINGER~\cite{adang2025singer} and GRaD-Nav++~\cite{chen2025grad} demonstrate language-conditioned drone navigation using learned visuomotor policies, but focus exclusively on navigation without physical manipulation or contact. More broadly, while cross-embodiment datasets such as Open X-Embodiment~\cite{open_x_embodiment_rt_x_2023} include dozens of manipulators, they contain no aerial platforms or underactuated flight dynamics, leaving the transferability of VLA representations to aerial manipulation unexplored.

% What are the key components of my approach and results? Also include any specific limitations. 
To address these gaps, we introduce \emph{AirVLA}, the first systematic investigation of whether manipulation-pretrained VLA foundation models can transfer to aerial manipulators through fine-tuning. Our system consists of a ModalAI Starling~2 Max quadrotor equipped with a lightweight compliant gripper derived from the Universal Manipulation Interface (UMI)~\cite{chi2024universal,ha2024umi, gupta2025umi}, forward- and downward-facing onboard cameras, and a ROS-based control stack that treats the drone as a ``flying end-effector.'' We construct a diagnostic task suite spanning (i) \emph{pick-and-place} manipulation collected via human teleoperation, (ii) \emph{navigation} tasks with human teleoperation data and synthetic training trajectories generated using model predictive control (MPC) in Gaussian Splatting reconstructions, and (iii) \emph{compositional} tasks that chain both behaviors. This design enables us to isolate which aspects of manipulation skill transfer to aerial platforms and which fail.

We directly fine-tune the foundation-scale VLA policy $\pi_0$ (pretrained on large manipulation datasets containing no aerial platforms) on our aerial manipulation data. Our experiments reveal partial but significant skill transfer: the fine-tuned policy learns stable hovering and coarse manipulation behaviors, demonstrating that foundation models capture some manipulation primitives that generalize to flight. However, standard fine-tuning fails at task completion due to sensitivity to payload dynamics and explicit obstacle navigation limitations. To address these specific failure modes, we introduce domain-adapted components: (i) a physics-aware low-level controller that wraps VLA outputs to enforce dynamically feasible commands, and (ii) synthetic navigation augmentation that enriches the training distribution. While these adaptations improve performance on navigation and stabilization during manipulation, fundamental limitations remain for long-horizon compositional manipulation and out-of-distribution adaptation, highlighting important open challenges in extending VLA models to underactuated, dynamic embodiments.

In summary, this paper makes the following contributions:
\begin{itemize}
    \item We present AirVLA, the first demonstration of a pre-trained VLA fine-tuned for an aerial manipulation platform.
    \item We manually collect a dataset of 270 aerial manipulation and navigation teleop-demos, to be open-sourced. 
    \item We synthesize 50 3DGS-based corrective navigation examples to help supervise VLA fine-tuning, showing a 20\% increase in navigation task success compared to fine-tuning with teleop only.
    \item We introduce a physics-informed guidance within real-time-chunking to adapt at runtime to the mass of a grasped object, improving task success by $10\%-40\%$ in pick and place tasks.
\end{itemize}

\section{Related Work}\label{sec:relatedwork}

\subsection{Vision-Language-Action Policies for Manipulation}
Vision-language-action (VLA) policies leverage large-scale multimodal pretraining to enable general-purpose robot control from natural language instructions. Early approaches such as RT-1~\cite{rt12022arxiv} and RT-2~\cite{rt22023arxiv} demonstrated that vision-language models pretrained on internet-scale data can be fine-tuned for robotic manipulation, mapping visual observations and language to action sequences. Subsequent work has scaled this paradigm through cross-embodiment training: the Open X-Embodiment dataset~\cite{open_x_embodiment_rt_x_2023} aggregates demonstrations from dozens of robot morphologies, enabling models like RT-X~\cite{open_x_embodiment_rt_x_2023} to generalize across diverse manipulators. More recent robot foundation models~\cite{bjorck2025gr00t,kim2024openvla} such as Octo~\cite{octo_2023} and $\pi_0$~\cite{pi-zero_2024} incorporate flow-matching and diffusion architectures, achieving state-of-the-art performance on manipulation benchmarks spanning tabletop tasks, mobile manipulation, and dexterous control.

These successes demonstrate that VLA policies can learn manipulation primitives that transfer across embodiments with different kinematics, workspaces, and end-effector designs. However, essentially all demonstrations in existing cross-embodiment datasets operate from stable bases (fixed or mobile) in quasi-static regimes. No prior work has investigated whether these learned representations transfer to aerial platforms, where underactuated flight dynamics, large ego-motion, and contact-induced disturbances fundamentally differ from ground-robot scenarios. Our work provides the first systematic evaluation of this transfer gap.

\subsection{Vision-Language and Vision-Language-Action Models for UAV Navigation}
Vision-language models have recently been applied to drone navigation~\cite{zhang2025grounded}, typically for high-level semantic reasoning rather than low-level visuomotor control. VLMaps~\cite{huang2022visual} constructs semantic maps for language-conditioned robot navigation but act as a perception and planning front-end rather than an end-to-end policy. SEEK~\cite{ginting2024seek} and VISTA~\cite{nagami2025vista} demonstrate uncertainty-aware exploration from semantic goals on quadrupeds and drones, but are limited to object localization and do not extend to interaction or contact. More recently, SINGER~\cite{adang2025singer} introduces an end-to-end visuomotor policy for language-conditioned drone navigation trained on synthetic trajectories in Gaussian Splatting environments, achieving onboard inference through semantic image preprocessing with CLIPSeg~\cite{luddecke2022image}. GRaD-Nav++~\cite{chen2025grad} extends this approach with differentiable dynamics in Gaussian Splatting scenes, enabling multi-task generalization across navigation behaviors. Both methods demonstrate impressive sim-to-real transfer for navigation tasks.

In contrast to SINGER and GRaD-Nav++, which focus on language-conditioned navigation without contact and use Gaussian Splatting primarily as a training environment, our work targets aerial manipulation with an onboard gripper, uses Gaussian Splatting as a source of synthetic demonstration data for a specific navigation subtask, and explicitly studies how a foundation VLA policy can be leveraged when navigation and manipulation must be composed in a single language-specified behavior.

\subsection{Aerial Manipulation with Learned Policies}
Unmanned aerial manipulation (UAM) combines multirotor flight with contact-rich interaction, introducing challenges such as underactuated dynamics, aerodynamic disturbances near surfaces, and strict payload limits. Classical systems demonstrate contact-based tasks including surface inspection~\cite{trujillo2019novel,bodie2019omnidirectional}, painting~\cite{8379422,guo2024flying,lanegger2022aerial}, drilling~\cite{ding2021design}, and object grasping~\cite{ghadiok2011autonomous, zhang2019compliant}, typically using carefully engineered controllers tailored to specific scenarios. Recent work has explored learning-based approaches: model-free reinforcement learning for aerial manipulation~\cite{cuniato2023learning}, one-shot learning for autonomous grasping~\cite{zito2022one}, and image-based visual servoing~\cite{6907149, 7759263}. However, these methods either lack language grounding or are limited to single tasks without generalization across manipulation primitives.

Most relevant to our work, UMI-on-Air~\cite{gupta2025umi} addresses the challenge of transferring manipulation policies to aerial platforms by training a single-task diffusion policy on handheld UMI demonstrations and introducing embodiment-aware guidance at inference time: gradient feedback from a low-level MPC controller steers trajectory generation toward dynamically feasible actions during deployment. This approach achieves impressive results on high-precision aerial manipulation tasks.

Compared to UMI-on-Air, which studies a single-task diffusion policy trained from handheld UMI demonstrations and introduces embodiment-aware guidance at inference time via MPC gradients, AirVLA studies cross-embodiment transfer in a generalist VLA foundation model ($\pi_0$) fine-tuned on aerial manipulation data. While both methods use guidance at inference time, AirVLA evaluates a broader set of tasks spanning manipulation, navigation, and the zero-shot composition of both. This design lets us isolate and study the representational generalization limits of existing VLA models for aerial manipulation and navigation.

More broadly, while prior aerial manipulation work either (i) relies on task-specific controllers without language grounding or (ii) uses cross-embodiment guidance to make existing policies feasible on drones, none of these efforts evaluate a generalist VLA manipulation model on a unified suite of pick-and-place, navigation, and compositional aerial tasks the way AirVLA does.
\section{Method}\label{sec:method}

\subsection{System Overview}

AirVLA is a vision-language-action system for aerial manipulation that transfers manipulation capabilities from foundation models pretrained on fixed-base robot arms to an underactuated quadrotor platform. The system takes as input RGB images from multiple viewpoints and a natural language task description, and outputs relative end-effector pose commands executed by a low-level flight controller.

The key challenge in this transfer is the mismatch between the quasi-static regimes of tabletop manipulation and aerial manipulation, where the platform must continuously stabilize against gravity while simultaneously executing precise gripper motions. Grasping an object introduces a step change in effective mass that, if uncompensated, causes the drone to sag and potentially fail the task.

Our approach addresses this challenge through two main contributions:
(1) a physics-aware guidance mechanism that augments the pretrained policy's action generation process with payload-aware vertical compensation at inference time, and
(2) a Gaussian-splatting data pipeline that enables efficient collection and synthesis of diverse training trajectories from a small number of seed flights.
Together, these allow a VLA model pretrained on large-scale robot manipulation data to perform aerial pick-and-place and navigation with minimal drone-specific fine-tuning.

% Fig.~\ref{fig:system_overview} illustrates the complete system pipeline from observation to action execution.

% \begin{figure}[t]
%     \centering
%     \includegraphics[width=\linewidth]{figures/system_overview.png}
%     \caption{System overview of AirVLA. Multi-view RGB images and language instructions are processed by the $\pi_0$ policy~\cite{pi-zero_2024} to generate action chunks. During inference, physics-aware guidance biases sampling toward dynamically stable trajectories, compensating for payload-induced disturbances.}
%     \label{fig:system_overview}
% \end{figure}

\subsection{Hardware}

The system integrates the ModalAI Starling 2 Max drone with a customized Universal Manipulation Interface (UMI) gripper~\cite{chi2024universal} and multiple cameras to enable autonomous aerial manipulation. The Starling 2 Max is powered by the VOXL 2 companion computer (Qualcomm QRB5165). The customized UMI-style gripper is attached to the underside of the drone, enabling dynamic grasping. The system integrates one external camera and two onboard cameras (downward and forward-facing), providing RGB images at 5Hz.

\subsubsection{Gripper Design}

\begin{figure}[t]
    \centering
    \includegraphics[width=\linewidth]{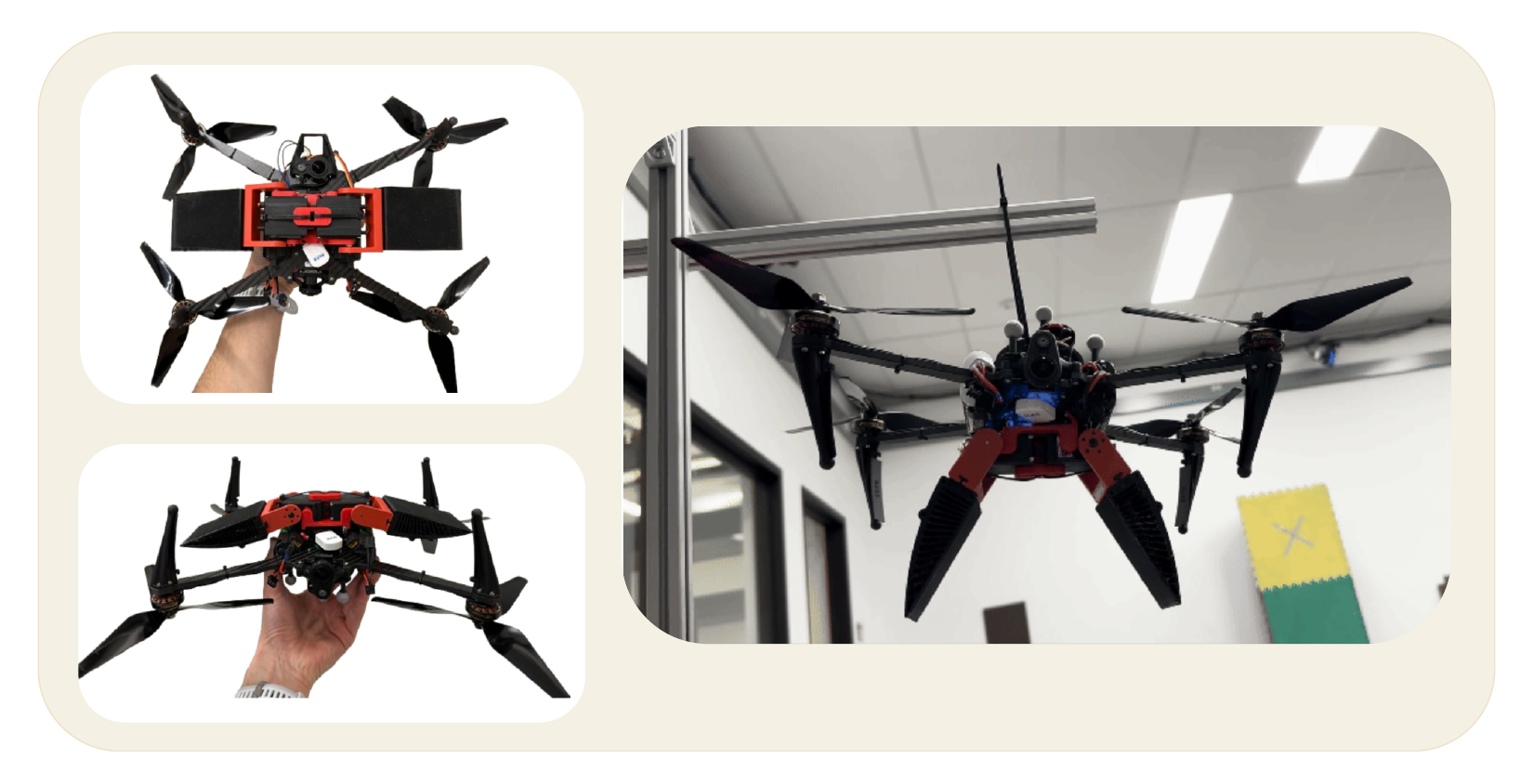}
    % \includesvg[width=\columnwidth]{figures/gripper/gripper.svg}
    \captionof{figure}{Custom gripper installed on the Starling 2 Max drone. The design prioritizes low weight for extended flight time while maintaining sufficient grip strength for the target objects.}
    \label{fig:gripper}
\end{figure}

We showcase the design of our gripper in Fig.~\ref{fig:gripper}. The gripper is designed to be built cheaply without specialized tools. The frame is entirely 3D printed, with two hobby-grade servos slotting in without screws. The fingers are adapted from the UMI gripper~\cite{chi2024universal}, facilitating direct comparison with arm-based policies trained with similar end-effector geometry.

\subsubsection{Observation and Action Spaces}

The observation space consists of RGB images from three cameras at $[256 \times 256]$ resolution, the drone's estimated pose from a motion capture system, and gripper aperture.
The action space, consisting of the drone position and yaw, is represented as an action chunk
\(
A \in \mathbb{R}^{H \times D}
\)
(i.e., end-effector 4 DoF delta poses and gripper commands $D$ over horizon \(H\)).
Actions are generated at 10 Hz and executed by the PX4 flight controller via position setpoints.

\subsection{Policy Architecture:}

We build on $\pi_0$~\cite{pi-zero_2024}, a vision-language-action model that represents the conditional action distribution using a flow-matching (continuous-time) generative model~\cite{lipman2023flowmatchinggenerativemodeling}. Given an observation \(o\) (images, proprioception, language), $\pi_0$ defines a velocity field
\(
v_\theta(x_\tau, o, \tau)
\)
over latent action chunks \(x_\tau \in \mathbb{R}^{H \times D}\) and diffusion/flow time \(\tau \in [0,1]\).
Sampling draws \(x_0 \sim \mathcal{N}(0, I)\) and integrates the ODE
\begin{equation}
    \frac{d x_\tau}{d\tau} = v_\theta(x_\tau, o, \tau)
    \label{eq:flow-ode}
\end{equation}
to obtain the generated action chunk \(A = x_1\).

For real-time execution, we employ Real-Time Chunking (RTC)~\cite{black2025real}, which enables asynchronous inference by \emph{freezing} the prefix of the next chunk that will execute before inference completes, and \emph{inpainting} the remaining suffix conditioned on the frozen prefix. Concretely, RTC defines a soft temporal mask over the horizon that blends previously committed actions with newly generated actions to avoid discontinuities at chunk boundaries.

\subsection{Physics-Aware Guidance for Action Generation}\label{sec:physics_guidance}

RTC shows that inference-time steering can be implemented by modifying the velocity field during sampling~\cite{black2025real}. We generalize this idea by introducing a loss \(\Phi(A; o)\) over the \emph{denoised} action chunk \(A = x_1\), and adding a gradient correction term derived from \(\nabla_A \Phi\) to the base velocity field, analogous in spirit to gradient-based guidance methods in generative modeling~\cite{ho2022classifierfreediffusionguidance,song2023pseudoinverse}.
Intuitively, we seek samples that both (i) have high probability under the base policy and (ii) minimize the guidance loss:
\begin{equation}
    p_{\text{guid}}(A \mid o) \propto p_\theta(A \mid o)\,\exp\!\bigl(-\Phi(A; o)\bigr).
    \label{eq:guided-density}
\end{equation}
This formulation seeks a 'sweet spot' between the policy's priors and physical constraints. It biases the sampling process toward actions that are high-probability under the VLA (preserving learned manipulation skills) while simultaneously minimizing the cost $\Phi$ (enforcing flight feasibility), effectively steering the drone at runtime without retraining.

Operationally, at intermediate time \(\tau\), we compute the model's current prediction of the terminal action chunk
\(
\hat{A}_\theta(x_\tau,o,\tau) \approx x_1
\)
(e.g., via the model's internal denoised estimate), evaluate the gradient \(\nabla_A \Phi(\hat{A}_\theta; o)\), and map it back to a correction in latent space through a vector--Jacobian product $\xi \coloneqq \Bigl(\nabla_{x_\tau}\hat{A}_\theta(x_\tau,o,\tau)\Bigr)^\top
\nabla_A \Phi\!\left(\hat{A}_\theta(x_\tau,o,\tau); o\right)$.
This yields the guided velocity field
\begin{equation}
v_{\text{guid}}(x_\tau,o,\tau)
=
v_\theta(x_\tau,o,\tau) - s(\tau)\xi.
\label{eq:guided-velocity}
\end{equation}

where \(s(\tau)\) is a scalar guidance schedule. Multiple loss terms compose additively because \(-\log p_{\text{guid}}\) adds losses, so RTC-style continuity objectives and task-specific physics objectives can be applied simultaneously without changing the pretrained model weights.

\subsubsection{General Tracking-Error Guidance}

Suppose we have a reference trajectory \(A^{\text{des}}(o) \in \mathbb{R}^{H\times D}\) derived from the current observation. We define a tracking loss
\begin{equation}
    \Phi_{\text{track}}(A; o)
    =
    \frac{1}{2}
    \sum_{t=0}^{H-1}\sum_{d=1}^{D}
    \lambda_d\, w_t\,
    \bigl(A_{t,d} - A^{\text{des}}_{t,d}(o)\bigr)^2,
    \label{eq:tracking-potential}
\end{equation}
with per-dimension strengths \(\lambda_d \ge 0\) and temporal weights \(w_t \ge 0\). The temporal weights follow the soft masking schedule from RTC \cite{black2025real}: \(w_t = 1\) for frozen prefix actions, the weights then exponentially decay for the intermediate region and are zero for freshly generated actions. This ensures the guidance strongly enforces continuity with committed actions while allowing freedom for new generation. The corresponding gradient correction is derived from
\begin{equation}
    \frac{\partial \Phi_{\text{track}}}{\partial A_{t,d}}
    =
    \lambda_d\, w_t\,\bigl(A_{t,d} - A^{\text{des}}_{t,d}(o)\bigr),
    \label{eq:tracking-grad}
\end{equation}
which pulls the denoised chunk toward the reference trajectory in each dimension proportionally to \(\lambda_d\).

\subsubsection{Payload-Aware Vertical Guidance}

In aerial manipulation, the dominant disturbance during grasping is an effective mass increase that manifests as vertical sag under load. Rather than modeling full 6-DOF dynamics, we instantiate~\eqref{eq:tracking-potential} only on the altitude-related action dimension \(d_z\):
\begin{equation}
    \Phi_{\text{payload}}(A; o, A_{t-1})
    =
    \frac{\lambda_z}{2}\,\alpha(o,A_{t-1})
    \sum_{t=0}^{H-1} w_t\,
    \bigl(z_t(A) - z_{\text{des}}(o)\bigr)^2,
    \label{eq:payload-potential}
\end{equation}
where \(z_t(A)\) denotes the \(z\)-component of the action at timestep \(t\), and
\begin{equation}
    z_{\text{des}}(o) = z_{\text{curr}}(o) + \Delta z
    \label{eq:zdes}
\end{equation}
biases the drone toward slightly higher altitude under load (\(\Delta z>0\)). Here \(z_{\text{curr}}(o)\) comes from a motion capture system, and \(\Delta z\) is a tuned offset capturing the expected sag for typical payloads.

The payload confidence \(\alpha(o,A_{t-1})\in[0,1]\) is computed from (i) the previously executed action chunk \(A_{t-1}\) and (ii) the current measured gripper aperture in \(o\). Concretely, we compute a smooth, bounded payload confidence by combining recent gripper command intent with the current measured aperture. Let $u_{t}\in[-1,1]$ denote the gripper command at timestep $t$ (with $+1$ corresponding to ``close''), and let $g(o)\in[0,1]$ denote the measured aperture (normalized so that larger values are more open). From the previously executed chunk $A_{t-1}$, we form continuous close/open intents using the last $K$ commands,
\begin{equation}
\begin{aligned}
c_{\text{intent}} &= \frac{1}{K}\sum_{i=H-K}^{H-1}\mathrm{clip}\!\left(\frac{u_i+1}{2},\,0,\,1\right), \\
o_{\text{intent}} &= \frac{1}{K}\sum_{i=H-K}^{H-1}\mathrm{clip}\!\left(1-\frac{u_i+1}{2},\,0,\,1\right),
\end{aligned}
\end{equation}
and compute soft aperture-based scores $c_{\text{meas}}, o_{\text{meas}}\in[0,1]$ from $g(o)$ (higher when the gripper is closed/open, respectively). We then define an ``open'' gate and payload confidence as
\begin{equation}
\begin{aligned}
o_{\text{flag}} &= \mathrm{clip}\!\left(\tfrac{1}{2}o_{\text{intent}} + \tfrac{1}{2}o_{\text{meas}},\,0,\,1\right), \\
\alpha(o,A_{t-1}) &= \mathrm{clip}\!\left((1-o_{\text{flag}})\,c_{\text{intent}}\,c_{\text{meas}},\,0,\,1\right).
\end{aligned}
\label{eq:alpha}
\end{equation}

When \(\alpha\approx 0\), the loss vanishes and sampling reduces to vanilla RTC; when \(\alpha\approx 1\), the sampler prefers chunks whose vertical component is biased toward \(z_{\text{des}}\), compensating for payload. This can be viewed as mode-dependent feedforward (gain scheduling / gravity compensation), but injected \emph{inside} the generative policy's sampling dynamics by adding the gradient correction term derived from \(\nabla_A \Phi_{\text{payload}}\) via~\eqref{eq:guided-velocity}.

\subsection{Gaussian Splat Data Pipeline}\label{sec:data_pipeline}

Collecting aerial manipulation demonstrations is time-consuming and requires skilled pilots. To efficiently generate diverse training data, we adopt a ``Flying in Gaussian Splats''-style approach that couples photorealistic Gaussian-splatting reconstructions with a lightweight drone dynamics model to synthesize large volumes of training data from a small set of seed demonstrations~\cite{10937041}. Concretely, our pipeline proceeds by first reconstructing the static environment as a Gaussian Splat and isolating the gripper visuals to prevent observation bias. We then couple a drone dynamics model with these assets to synthesize diverse, physics-feasible training trajectories that cover both nominal navigation and recovery behaviors.

\subsubsection{Scene Reconstruction}

We reconstruct each scene from short walk-throughs captured with the drone's forward-facing camera. This results in a set of metrically scaled poses for each image, which are used to train a 3D Gaussian splatting model~\cite{kerbl20233d} using Nerfstudio~\cite{tancik2023nerfstudio,ye2025gsplat}. The resulting model \(\mathcal{GS}_\phi\) renders photorealistic images from arbitrary camera poses in the captured region. Given a camera pose \((p,q)\), the rendered image is
\(
I=\mathcal{GS}_\phi(p,q).
\)

\subsubsection{Gripper Segmentation and Compositing}

The downward-facing camera provides critical visual information for manipulation, but the gripper is persistently visible in its field of view. Including raw downward facing images in the synthetic training data from the Gaussian splatting model would introduce an observation bias into the policy that would result in unwanted behavior. Thus, we explicitly treat the gripper as a separate foreground layer and composite it onto renders from a gripper-free scene model. Concretely, for each downward-facing training frame \(I_{\text{down}}\), we compute a gripper mask \(M_{\text{grip}}\) using Segment Anything (SAM)~\cite{kirillov2023segment,ren2024grounded} with a fixed bounding box prompt corresponding to the gripper's known image-space location. We then extract a masked gripper patch
\begin{equation}
    G = I_{\text{down}} \odot M_{\text{grip}},
\end{equation}
to form a small library \(\{G_a\}\) of representative gripper appearances.

We train the Gaussian splatting model \(\mathcal{GS}_\phi\) on images from cameras where the gripper is not visible (namely, the forward-facing views). At synthesis time, we render the clean scene from \(\mathcal{GS}_\phi\) for the downward-facing viewpoint and composite the gripper foreground:
\begin{equation}
    I_{\text{down}}^{\text{synth}}
    =
    (1-M_{\text{grip}})\odot \mathcal{GS}_\phi(p,q)
    \;+\;
    M_{\text{grip}}\odot G_{a(t)}.
\end{equation}
This yields a downward-view image whose background is fully determined by the scene model, while the gripper appearance is consistent with the commanded aperture \(a(t)\). 

% Importantly, we do not attempt to hallucinate or reconstruct the background behind the gripper; the masked region is always occupied by the composited gripper layer.

% \begin{figure}[t]
%     \centering
%     \includegraphics[width=\linewidth]{figures/inpainting_pipeline.png}
%     \caption{Gripper compositing pipeline. (a) Raw downward-facing image with gripper visible. (b) SAM segmentation mask. (c) Clean downward-view render from the Gaussian splat scene model. (d) At render time, we composite a state-appropriate gripper foreground onto the rendered image.}
%     \label{fig:inpainting_pipeline}
% \end{figure}

\subsubsection{Drone Dynamics Model}

Following~\cite{10937041}, we simulate drone motion with a semi-kinematic state \(x=(p^W,v^W,q^W_B)\) comprising position, velocity, and orientation quaternion. Controls are normalized thrust \(f_{\text{th}}\) and body angular velocity \(\omega^B\). The dynamics are
\begin{align}
    \dot{p}^W &= v^W, \\
    \dot{v}^W &= g\,e_3^W + \frac{k_{\text{th}}}{m}\, f_{\text{th}}\, R(q^W_B)\, e_3^B, \\
    \dot{q}^W_B &= \tfrac{1}{2}\,\Omega(\omega^B)\, q^W_B,
\end{align}
where \(g\) is gravitational acceleration, \(e_3^W,e_3^B\) are the \(z\)-axis unit vectors in world/body frames, \(R(\cdot)\) converts quaternion to rotation, and \((k_{\text{th}},m)\) are thrust coefficient and mass. We forward-integrate using ACADOS~\cite{verschueren2022acados} and render images using the body-to-camera transform \(T^B_C\).

\subsubsection{Domain-Randomized Data Synthesis}

To enable the policy to recover from dangerous states near the obstacle, we additionally synthesize recovery trajectories by randomizing task geometry, similar to FiGS~\cite{10937041}. For each nominal trajectory \((X^d,U^d)\), we generate \(N_s\) randomized rollouts by sampling an initial state perturbation \(x^s_0 \sim \mathcal{U}(x^d_0-\Delta x,\,x^d_0+\Delta x)\), then simulating the resulting trajectory rendering multi-view images from the Gaussian splat scene model \(\mathcal{GS}_\phi\).

We additionally randomize intermediate waypoints to increase diversity and induce recovery behaviors. For the navigation task, we randomize the terminal hover location to lie above the goal by sampling a height offset uniformly in \([1.0, 1.5]\) meters. Let \(p_{\text{obj}}\) denote the object goal position in the scene and let \(h \sim \mathcal{U}(1.0, 1.5)\,\text{m}\). We set the goal position as
\begin{equation}
    p_{\text{goal}} = p_{\text{obj}} + [0,0,h]^\top.
    \label{eq:goal-rand}
\end{equation}

To diversify gate exits, we randomize the post-gate waypoint within a ball of diameter \(0.25\) m centered at a nominal after-gate position waypoint \(p_{\text{after}}^d\). With radius \(r = 0.125\) m and \(\delta \sim \mathcal{U}(\mathbb{B}(0,r))\) where \(\mathbb{B}(0,r)=\{\delta\in\mathbb{R}^3:\|\delta\|_2\le r\}\), we define
\begin{equation}
    p_{\text{after}} = p_{\text{after}}^d + \delta.
    \label{eq:aftergate-rand}
\end{equation}

Finally, we perturb trajectories to elicit recovery behavior by inserting an additional waypoint between the start and the gate that forces the drone to pass near one of four gate extremities, namely top, bottom, left, or right. We sample a side \(s \sim \mathrm{Unif}(\{\text{top},\text{bottom},\text{left},\text{right}\})\) and define a side-specific waypoint
\begin{equation}
    p_{\text{wp}} = \bar{p}_s + b\,n_s,
    \label{eq:recovery-wp}
\end{equation}
where \(\bar{p}_s\) is a reference point near the corresponding gate boundary, \(n_s\) is the offset direction away from the gate frame, and \(b>0\) is a fixed geometry buffer chosen to ensure collision-free clearance.

\newcommand{\trajH}{1.15in}

\begin{figure}[t]
  \centering

  %\begin{minipage}[b]{0.49\columnwidth}
    %\centering
    %\includegraphics[height=\trajH,keepaspectratio]{figures/gs_data/left_gate_nominal.png}
    %\vspace{-1.2mm}
    %{\scriptsize (a) Left gate, nominal}
  %\end{minipage}\hfill
  %\begin{minipage}[b]{0.49\columnwidth}
    %\centering
    %\includegraphics[height=\trajH,keepaspectratio]{figures/gs_data/right_gate_nominal.png}
    %\vspace{-1.2mm}
    %{\scriptsize (b) Right gate, nominal}
  %\end{minipage}

  %\vspace{1mm}

  %\begin{minipage}[b]{0.49\columnwidth}
    %\centering
    %\includegraphics[height=\trajH,keepaspectratio]{figures/gs_data/left_gate_corrective_angle3.png}
    %\vspace{-1.2mm}
    %{\scriptsize (c) Left gate, corrective}
  %\end{minipage}\hfill
  %\begin{minipage}[b]{0.49\columnwidth}
    %\centering
    %\includegraphics[height=\trajH,keepaspectratio]{figures/gs_data/right_gate_corrective_angle2.png}
    %\vspace{-1.2mm}
    %{\scriptsize (d) Right gate, corrective}
  %\end{minipage}
  % \includesvg[width=\columnwidth]{figures/gs_data/gs_data_combined_labels.svg}
  \includegraphics[width=\columnwidth]{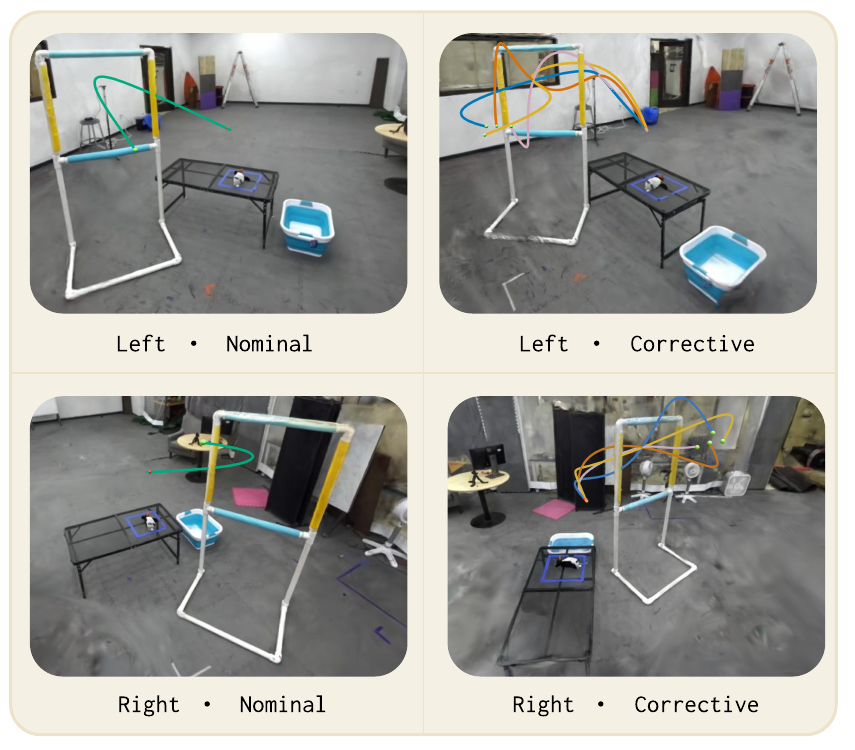}
  % \includesvg[width=\columnwidth]{figures/gs_data/gs_data_combined_no_center_labels.svg}

  \vspace{-1mm}

  \caption{Examples of synthetic trajectories used to generate training data. Nominal rollouts and corrective rollouts overlay sampled trajectories in the gate-manipulation scene.}
  \label{fig:synth_traj_examples}
\end{figure}

After sampling these randomized waypoint targets, we generate controls by tracking the waypoint-conditioned plan under the drone dynamics, producing the rollout \((X^s,U^s)\). Figure \ref{fig:synth_traj_examples} depicts nominal and corrective synthetic trajectories for both the left gate and right gate scenarios. We render images along \(X^s\) from \(\mathcal{GS}_\phi\) using the body-to-camera transform \(T^B_C\), and composite the gripper for downward-facing views. This procedure yields a large set of physically and geometrically plausible trajectories that cover both nominal executions and off-nominal approaches requiring recovery near the gate.

\section{Experiments}\label{sec:experiments}
\begin{table*}[!t]
\centering
\caption{\textbf{Single Task Performance Benchmarks (\%).} Each method was evaluated across twenty trials per task.}
\label{tab:modular_results}
\small
\begin{tabular}{l cc @{\hspace{1.5em}} cc @{\hspace{1.5em}} cc}
\toprule
\multirow{2}{*}{\textbf{Method}} & \multicolumn{2}{c}{\textbf{Penguin Grasp}} & \multicolumn{2}{c}{\textbf{Navigation (Non-Synthetic)}} & \multicolumn{2}{c}{\textbf{Navigation (Synthetic)}} \\
\cmidrule(r){2-3} \cmidrule(r){4-5} \cmidrule{6-7}
  & Pick & Place & Gate & Hover & Gate & Hover \\
\midrule
$\pi_0$  naive                      & 50.0 & 0.0  & 50.0 & 60.0 & 45.0 & 100.0 \\
$\pi_0$ + RTC                       & 85.0 & 23.5 & 80.0 & 81.2 & 95.0 & 100.0 \\
$\pi_0$ + RTC w/ payload-aware guidance  (ours)   & 100.0& 50.0 & --   & --   & --   & --    \\
ACT                                 & 0.0  & 0.0  & 0.0  & 0.0  & 0.0  & 0.0   \\
Diffusion Policy                    & 10.0 & 0.0  & 15.0 & 0.0  & 0.0  & 0.0   \\
\bottomrule
\end{tabular}
\end{table*}
To evaluate our methods, we define a task-specific score rubric, report quantitative results under that rubric, and organize experiments to isolate the contribution of pretraining transfer, inference-time stabilization, and data augmentation.

Through our experiments, we seek to study the following questions:
\begin{enumerate}
    \item \textbf{Transfer:} How much manipulation-pretrained VLA capability transfers to an underactuated aerial manipulator after fine-tuning?
    \item \textbf{Inference-time control:} Can real-time chunking (RTC) and payload-aware guidance reduce the sensitivity of aerial execution to chunk discontinuities and payload disturbances?
    \item \textbf{Data synthesis:} Does Gaussian-splat-based synthetic navigation augmentation improve real-world navigation performance?
    \item \textbf{Compositionality:} Can a single policy reliably compose navigation and grasping into a multi-stage task (navigate-then-grasp)?
\end{enumerate}

\subsection{Task suite}
% QVjan30-RESOLVED: I don't think we need to explain the robot platform again here, actually, because we did quite an extensive job in the method section. It's also a little confusing to be describing the evaluation task suite under a robot platform paragraph
% \textbf{Robot platform.}
% Our system is a ModalAI Starling 2 Max quadrotor equipped with a lightweight compliant gripper and forward/downward RGB cameras, controlled via a ROS-based stack that treats the drone as a ``flying end-effector.'' We evaluate policy variations on a diagnostic task suite spanning (i) pick-and-place, (ii) navigation with and without synthetic data augmentation, and (iii) compositional tasks chaining both behaviors with and without synthetic data augmentation.
We evaluate using the following task prompts:
\begin{itemize}
    \item \textbf{Penguin Grasp (Pick-and-Place):} pick up the stuffed animal and put it in the blue bin.
    \item \textbf{Gate Navigation:} fly through the gate and hover over the stuffed animal.
    \item \textbf{Compositional (Navigate-then-Grasp):} fly through the gate and hover over the stuffed animal and then pick up the stuffed animal and put it in the blue bin.
\end{itemize}
The penguin grasp task evaluates the model's ability to transfer manipulation skills from pretraining to a drone embodiment. To ensure robustness, we vary the object's position within a box during both data collection and evaluation. The gate navigation task tests drone-specific navigation requiring explicit obstacle avoidance. By using two different gate positions (left and right), we force the policy to localize the gate prior to navigation. Finally, to test the model's ability to handle novel, composite tasks in a zero-shot manner, we fine-tune the policy on the combined datasets and evaluate it using a combined prompt unseen during training.

% Why did we choose these tasks in particular?
% We wanted to choose a representative manipulation task. It's 
% We also wanted to choose a task with explicit obstacle avoidance and navigation
% Finally we wanted to determine how well the model can combine tasks in a zero shot manner

\subsection{Evaluation rubric and protocol}
% QV jan 30: What is navigation (synthetic) in table one? Is an explanation of why we ran that particular evaluation that the navigation in the GS aims to show that it's a good sanity check before running navigation in real?
For each task, we define a rubric that measures progress toward completion. In our setting, the primary metric is binary task success, and we additionally record a diagnostic failure taxonomy to attribute failures for qualitative analysis:
\begin{itemize}
    \item \textbf{Penguin Grasp:} a two stage task with success markers triggered after (1) picking up the stuffed animal and (2) placing it in the bin. Failure modes include grasping followed by immediate drops or crashes as well as failure to pick. 
    \item \textbf{Gate Navigation:} a two stage task with success markers triggered after (1) navigating through the gate and (2) holding a hover over the stuffed animal. Failure modes include passing through the gate but not hovering over the penguin, and crashing or failing to pass through the gate.
    \item \textbf{Compositional Navigate and Grasp:} a four stage task combining the success and failure markers for the above with an additional failure marker of \emph{incorrect subtask order} (attempted grasp before passing the gate).
\end{itemize}

\subsection{Methods compared}
We compare the following methods:
\begin{itemize}
    \item \textbf{$\pi_0$ naive:} roll out action chunks, pausing for inference at the end of each action chunk.
    \item \textbf{$\pi_0$ + RTC:} real-time chunking that re-samples suffix actions to avoid discontinuities~\cite{black2025real}.
    \item \textbf{$\pi_0$ + RTC + payload-aware guidance (ours):} inference-time steering implemented via a gradient correction term applied inside the sampler dynamics~\cite{black2025real}.
    \item \textbf{ACT}~\cite{zhao2023learningfinegrainedbimanualmanipulation} and \textbf{Diffusion Policy}~\cite{chi2023diffusionpolicy}
\end{itemize}

For the navigation and compositional tasks, we additionally evaluate \textbf{synthetic} and \textbf{non-synthetic} variants, corresponding to whether the policy was trained with synthetic data augmentation. Training details for these methods are provided in Appendix VII-A.

\subsection{Quantitative results}
Each of the above methods are evaluated across twenty trials per task for a cumulative 460 flight trials. We omit evaluations of the payload-aware guidance method on pure navigation tasks, as the guidance method is identical to RTC when objects are not being manipulated. We present the results of these evaluations in Tables \ref{tab:modular_results} and \ref{tab:compositional_refined}. The reported success rates are conditioned on the number of trials that succeeded in the prior stage of the task. For example, the place success rate is conditioned on the number of trials that succeeded in picking the object.

To evaluate generalization to novel objects, we conduct 10 additional pick-and-place trials, replacing the objects used during training with previously unseen ones. Furthermore, to assess spatial robustness, we vary the gate location across three regions (front, left, and right). We conduct five trials per region, shifting and rotating the gate within each area. These out-of-distribution results are detailed in Table \ref{tab:ood_robustness_lean}. We discuss these results further in Appendix VII-B, and interpret them as evidence of partial semantic transfer together with substantial overfitting in the small-data fine-tuning regime for navigation.

\begin{table*}[t]
\centering
\caption{\textbf{Compositional Navigate-then-Grasp Success (\%).} Each method was evaluated across twenty trials per task.}
\label{tab:compositional_refined}
\small
\begin{tabular}{ll cccc}
\toprule
\textbf{Data Setting} & \textbf{Method} & \textbf{Gate} & \textbf{Hover} & \textbf{Pick} & \textbf{Place} \\
\midrule
\multirow{5}{*}{No Synthetic} & $\pi_0$   naive                   & 35.0 & 85.7 & 42.9 & 0.0 \\
                              & $\pi_0$ + RTC                     & 80.0 & 100.0 & 81.2 & 15.4 \\
                              & $\pi_0$ + RTC w/ payload-aware guidance (ours) & 70.0 & 100.0 & 100.0 & 35.7 \\
                              & ACT                               & 0.0  & 0.0  & 0.0  & 0.0 \\
                              & Diffusion Policy                  & 5.0  & 0.0  & 0.0  & 0.0 \\
\cmidrule{2-6}
\multirow{5}{*}{Synthetic}    & $\pi_0$ naive                     & 70.0 & 85.7 & 25.0 & 0.0 \\
                              & $\pi_0$ + RTC                     & 95.0 & 94.7 & 83.3 & 20.0 \\
                              & $\pi_0$ + RTC w/ payload-aware guidance (ours) & 85.0 & 100.0 & 94.1 & 62.5 \\
                              & ACT                               & 25.0 & 0.0  & 0.0  & 0.0 \\
                              & Diffusion Policy                  & 0.0  & 0.0  & 0.0  & 0.0 \\
\bottomrule
\end{tabular}
\end{table*}

\begin{table}[t]
\centering
\caption{\textbf{Out-of-Distribution (OOD) Robustness (\%).}Success rates for variations in object class and gate positioning. \emph{Place} and \emph{Hover} are reported as conditional percentages.}
\label{tab:ood_robustness_lean}
\small
\begin{tabular}{l cc @{\hspace{3em}} l cc}
\toprule
\multicolumn{3}{l}{\textbf{Grasp (Object Variation)}} & \multicolumn{3}{l}{\textbf{Navigation (Gate Locations)}} \\
\cmidrule(r){1-3} \cmidrule(l){4-6}
\textbf{Object} & \textbf{Pick} & \textbf{Place} & \textbf{Location} & \textbf{Gate} & \textbf{Hover} \\
\midrule
Chips           & 10.0          & 0.0           & Front             & 0.0          & -- \\
Sandwich        & 70.0          & 57.1           & Left       & 0.0          & -- \\
Box             & 30.0          & 33.3           & Right      & 40.0          & 100.0  \\
\bottomrule
\end{tabular}
\end{table}

\subsection{Analysis}
% Missing from the analysis:
% - 

\noindent\textbf{Inference-time structure is critical for aerial pick-and-place.
}Naive fine-tuning is insufficient for full pick-and-place success, attaining \(0 \%\), with failures dominated by missed grasps and post-contact disturbances. RTC improves success (up to \(23.5\%\)) by stabilizing execution across chunk boundaries, and payload-aware guidance further improves success (up to \(50\%\)), consistent with the need to compensate for payload-induced altitude sag and other underactuated dynamics. Although the pick-and-place task is deliberately constrained relative to unconstrained 6-DoF aerial manipulation, it remains challenging for end-to-end visuomotor learning, as reflected in the poor performance of ACT and Diffusion Policy. This suggests that embodiment-specific adaptation, such as payload-aware guidance, is necessary to achieve reliable success even in controlled top-down grasping due to the underactuated flight dynamics. 

\noindent\textbf{RTC substantially improves gate navigation.}
In the non-synthetic gate navigation trials, RTC increases gate traversal success from $50\%$ to $80\%$ and from $45\%$ to $95\%$ in the synthetic gate navigation trials. This indicates that RTC is not merely helpful but necessary for reliable navigation, since re-planning at runtime mitigates compounding errors during aggressive motion.

\noindent\textbf{Synthetic augmentation helps most when paired with RTC.}
In the synthetic setting, naive performance is asymmetric across gates while $\pi_0$+RTC achieves $95\%$ success at the gate traversal component of the task. This suggests that synthetic augmentation is valuable but insufficient on its own: it expands coverage of approaches and recoveries, while reliable deployment still depends on inference-time stabilization from RTC. Qualitatively, we observed that the policy would periodically fall back on "synthetic modes" that replay the GS-generated trajectories, ensuring successful gate traversal. 

\noindent\textbf{Compositional tasks reveal an ability to generalize from atomic tasks.}
Our method shows a strong ability to generalize to compositional tasks in a zero-shot matter attaining a \(62\%\) overall conditioned success rate. The compositional task evaluations further highlight the importance of closed-loop policy inference as well as the necessity of GS synthetic trajectories to ensure successful gate traversal.

\noindent\textbf{Out-of-distribution trials demonstrate generalization capabilities.} Our method generalizes effectively in pick-and-place tasks, achieving up to 57\% task success on a novel sandwich object. However, performance varies by object geometry: the model achieved only a 10\% pick success rate with a bag of chips, compared to 70\% for the sandwich. The primary failure mode was the policy's inability to adapt grasp poses to the chip bags geometry. In the navigation task, the policy adapts best to the "right" region (40\% success). Conversely, trials in the "front" and "left" regions failed completely. Specifically, the drone consistently bypassed the gate in front-region trials, while left-region trials resulted in collisions due to incomplete position adaptation.
% Is loss balancing necessary?
% How i
% 355, .15
\section{Limitations and Future Work}\label{sec:conclusion}
% In conclusion, we have presented the first steps towards a foundational vision language action model intended for aerial vehicles. The combinations of LLMs and aerial manipulators has the potential to transform the way humans utilize drones. The impressive maneuverability of drones allows them to reach places no other robot or human can, while the integration of VLA models enables unprecedented spatial reasoning capabilities. We hope that our novel UMI based drone gripper can be used across many platforms and research projects in the future. Furthermore our platform enables, along with its ROS base interface allows any roboticist to plug drone into their pipeline treating them simply as a robot arm end effector.  We collect demos on over 13 different tasks which we plan on open sourcing for use by the wider robotics community. Although our fine-tuned $\pi_0$ policy does not successfully complete its task, we do observe an emergent learned hovering behavior along with an attempted grasp. Moving forward, expect that collecting more demos will significantly improve the performance of the fine-tuned policy. Furthermore we plan to include the drones internal state estimation algorithm as an input to our policy. 

We present AirVLA, the first systematic evaluation of transferring a manipulation-pretrained VLA foundation model to an aerial platform. Our results provide a nuanced answer to the challenge of cross-embodiment transfer: while the visual and semantic representations of models like \(\pi_0\) transfer robustly to drone viewpoints, the intricate dynamics of aerial manipulation require specific interventions. By integrating payload-aware guidance directly into the flow-matching sampling loop, we improve single-task placement success from 0\% (naive baseline) to 50\%, reconciling the generative diversity of a VLA with the strict physical constraints of flight. Furthermore, we show that the policy can execute compositional tasks in a zero-shot manner, achieving 85\% gate traversal and up to 62.5\% place success rates. Although our method currently relies on motion capture to isolate the cross-embodiment transfer problem from localization engineering issues; future work will replace this dependency with onboard VIO or SLAM. Doing so would expand the operating domain to settings such as multi-room navigation that are difficult to study under motion capture. Finally, our out-of-distribution evaluations highlight a disparity in generalization: while the model achieves 57\% success on novel object classes, the lower 40\% success rate in novel navigation scenarios indicates that significantly more data is needed to enable robust aerial navigation.

% Limitations:
% - MOCAP but pipeline can be extended to VIO without any changes
% - OOD performance indicates the policy overfit to the training data. This indicates we need more drone-specific manipulation and navigation data to improve generalization
% - Future work should seek to extend the pipeline by using VIO in larger environments such as multi-room navigation and pick-and-place
% - 

% Furthermore, our diagnostic task suite reveals a critical dichotomy: while low-level manipulation and navigation skills were successfully acquired, the VLA fails to generalize to long-horizon compositional tasks. This failure mode suggests that future work must focus not only on control-level adaptation but also on preserving the high-level reasoning capabilities of VLAs during fine-tuning—perhaps through hierarchical architectures that separate semantic planning from dynamic execution. Ultimately, AirVLA establishes a strong baseline and a methodological blueprint for bringing the reasoning power of foundation models to the dynamic, three-dimensional world of aerial robotics.
\section{Acknowledgments}\label{sec:ack}
The authors would like to thank Timothy Chen and Ola Shorinwa for their advice and reviews in the paper writing stage.  This work was supported in part by a gift from Google, NSF grant FRR 2342246, and ONR grant N00014-23-1-2354.

%% Use plainnat to work nicely with natbib. 

\bibliographystyle{plainnat}
\bibliography{references}
\newpage

\begin{figure*}[!ht]
    \centering
    \includegraphics[width=0.30\textwidth,height=4cm,keepaspectratio]{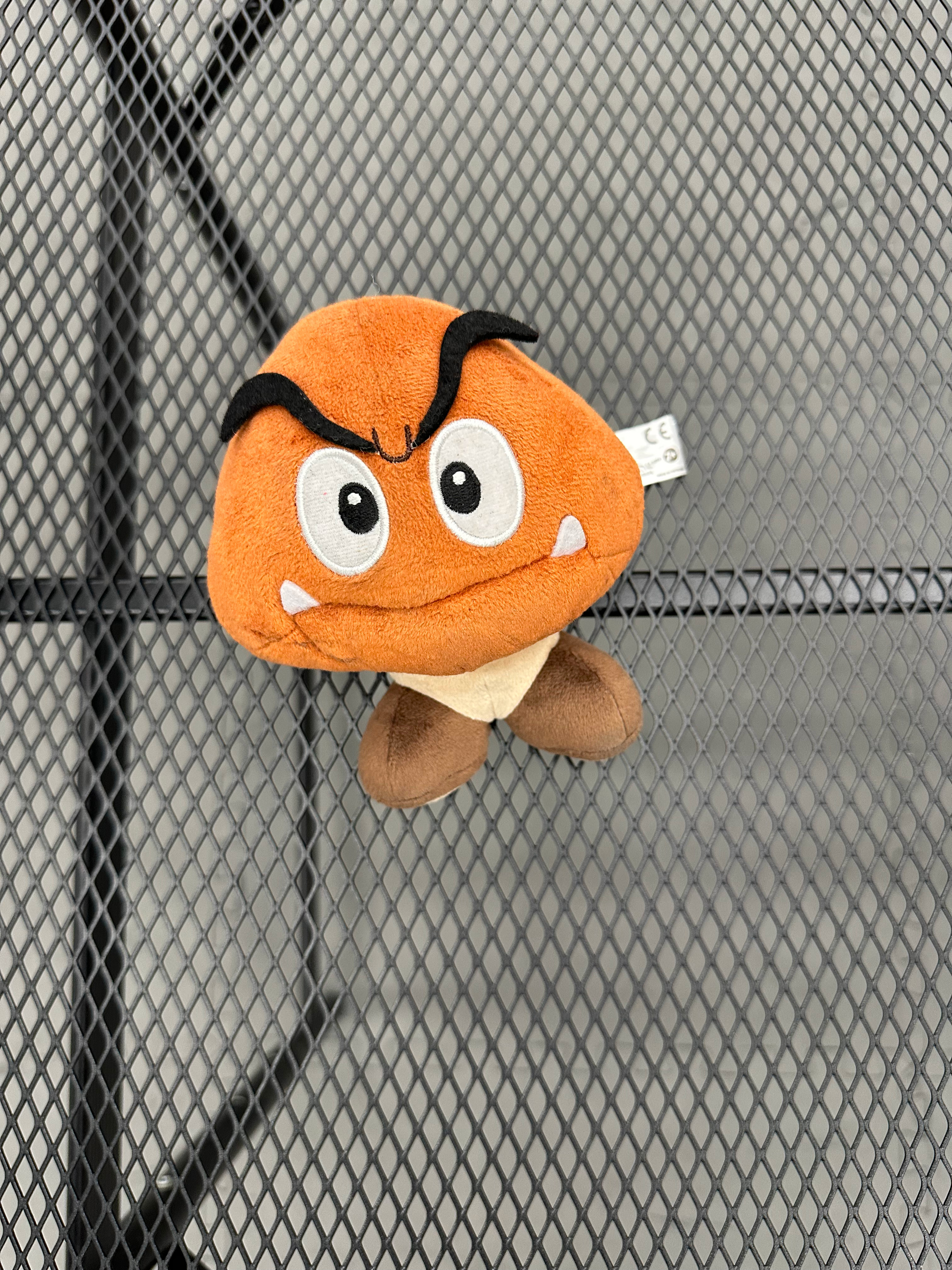}
    \includegraphics[width=0.30\textwidth,height=4cm,keepaspectratio]{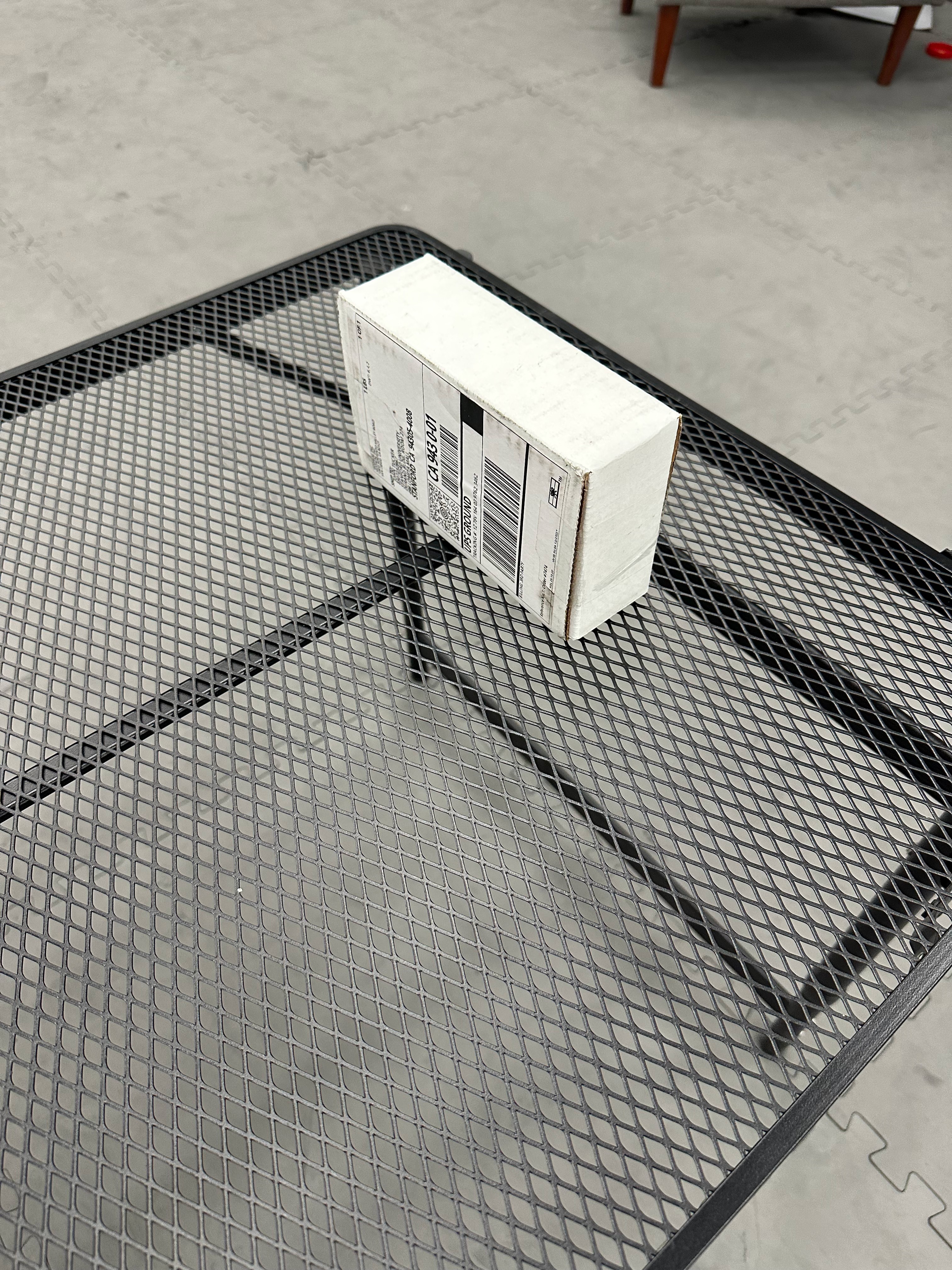}
    \includegraphics[width=0.30\textwidth,height=4cm,keepaspectratio]{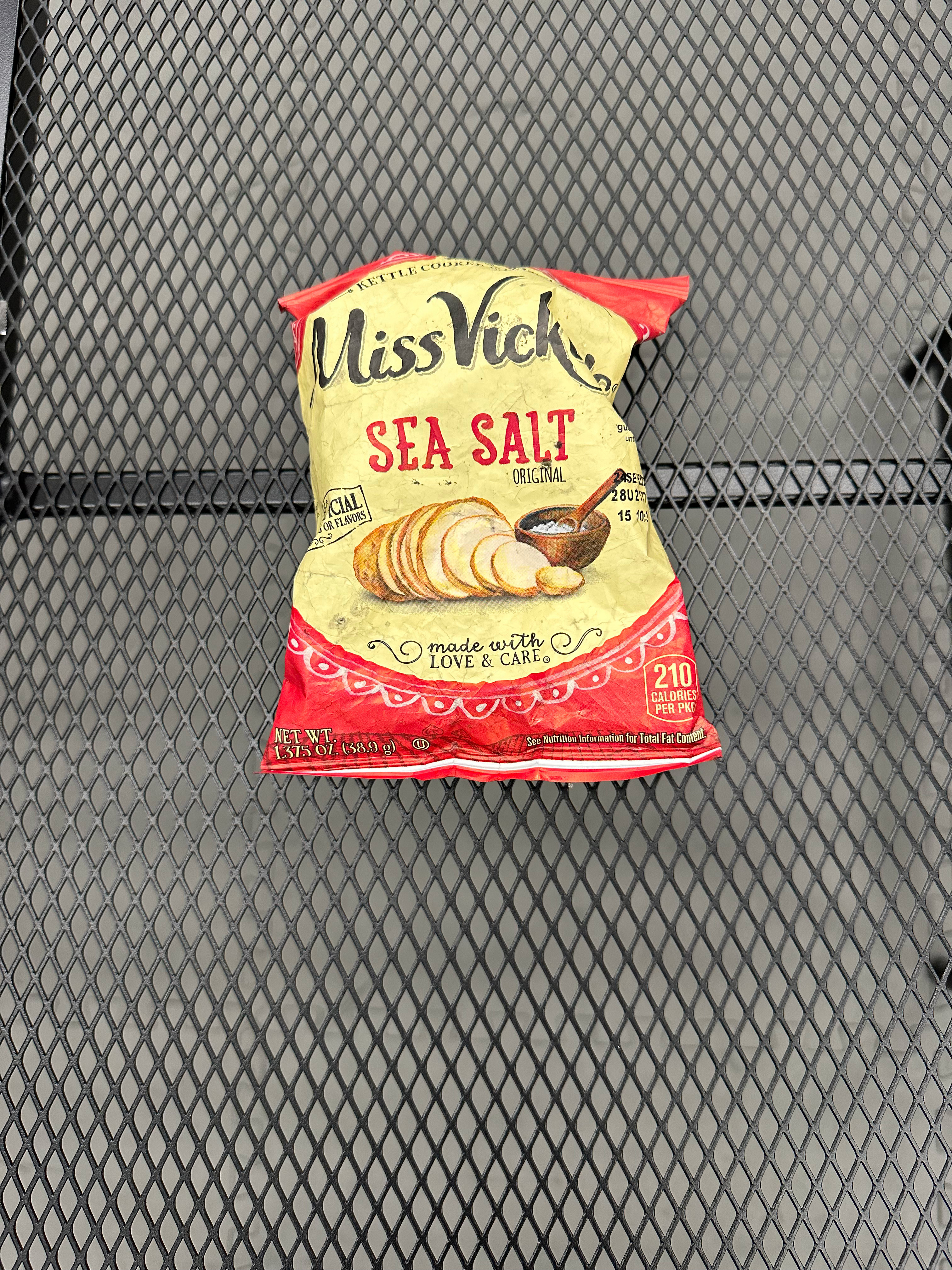}
    \includegraphics[width=0.30\textwidth,height=4cm,keepaspectratio]{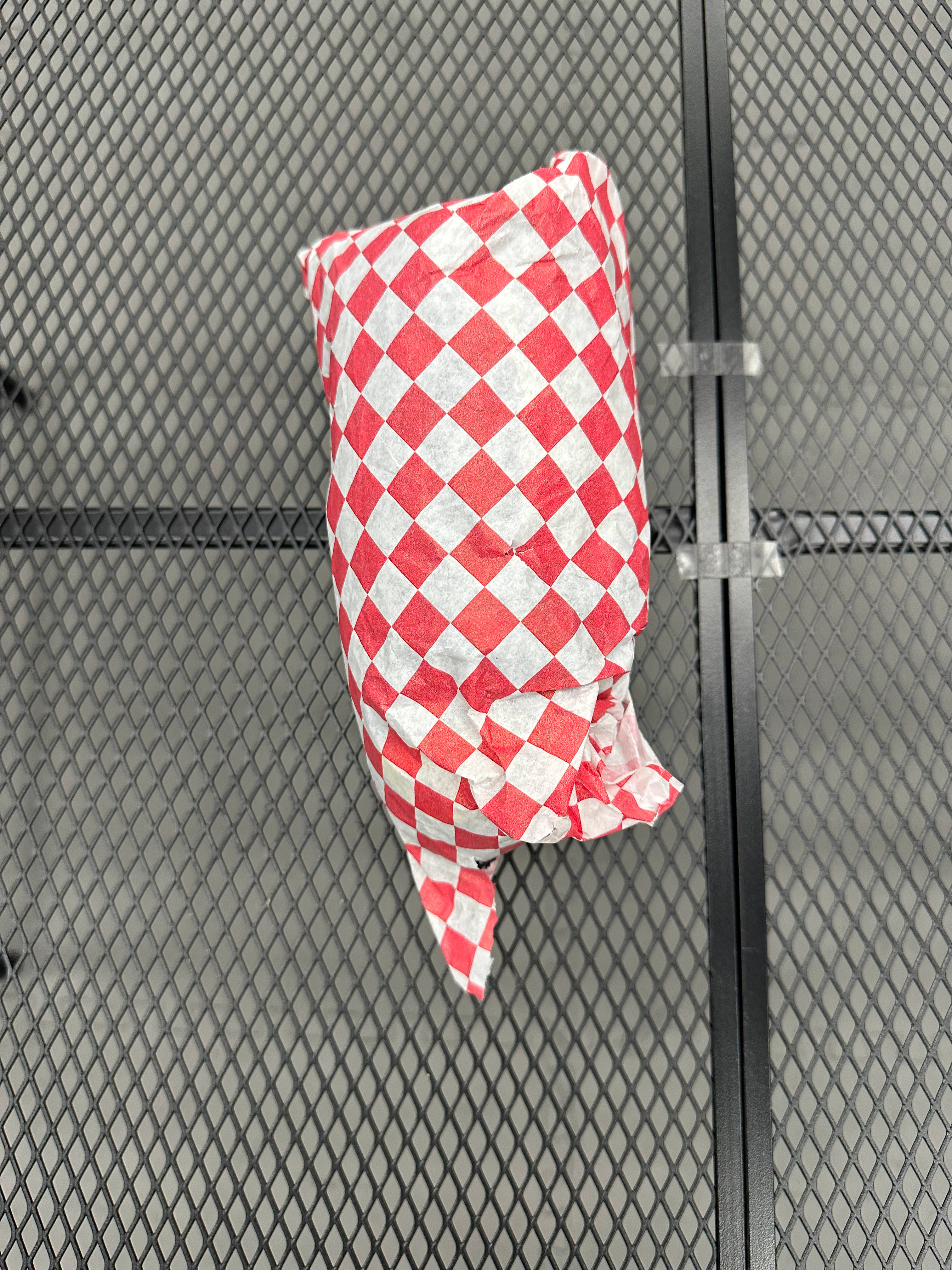}
    \caption{Out-of-distribution objects used for generalization probes. From left to right: goomba, box, chips, and sandwich. The policy was fine-tuned on a penguin but evaluated on these novel objects to test object-level generalization beyond training instances.}
    \label{fig:ood}
\end{figure*}

\section{Appendix}\label{sec:appendix}

\subsection{Training Details}\label{sec:training}
Here we provide training details for the fine-tuned $\pi_0$ policies. The baseline methods were trained using the LeRobot~\cite{cadene2024lerobot} code base with default configurations.

\subsubsection{Pretraining}

We initialize from the public $\pi_0$ base checkpoint~\cite{pi-zero_2024}, which is pretrained on large-scale multi-embodiment robot manipulation data.

\subsubsection{Fine-tuning}

We fine-tune on our drone dataset consisting of 120 to 150 (or about 10 hours) of real demonstrations for the manipulation and navigation tasks, respectively. The navigation dataset is further augmented to 200 total trajectories via Gaussian-splat trajectories (Sec.~\ref{sec:data_pipeline}). Fine-tuning is performed for $30{,}000$ gradient steps using AdamW optimization with a cosine learning-rate schedule with warmup: the learning rate linearly warms up over the first $1{,}000$ steps to a peak value of $2.5\times 10^{-5}$ and then decays following a cosine schedule to $2.5\times 10^{-6}$ by the end of training. We use a global batch size of $32$ and an action chunk horizon of $H=50$. Observations consist of three RGB views (one external, two body-mounted) and robot proprioception. Actions are represented in the dataset's native 4-DoF plus gripper control space. During training, we pad both proprioception and actions to match the model's expected dimensionality; at inference, we extract the first 7 action dimensions as the gripper is in the 7th dimension.

\subsubsection{Inference Configuration}

At inference time, for the vanilla implementation, actions are executed at $10\,\mathrm{Hz}$ and we run inference once every 50 executed actions (i.e., at $0.2\,\mathrm{Hz}$). We use RTC~\cite{black2025real} with a chunk horizon of $H=25$, using an exponential prefix-attention schedule with a prefix-attention horizon of 25 and $10$ denoising steps per chunk.

For payload-aware guidance (Sec.~\ref{sec:physics_guidance}), we set $\lambda_z=0.5$, $\Delta z=0.15\,\mathrm{m}$, and $w_t=(t/(H-1))^\gamma$ with $\gamma=1$, concentrating guidance on later timesteps where post-grasp transport dominates. The payload indicator used to gate the vertical objective (Eq.~\ref{eq:alpha}) is computed via a deterministic heuristic from the previous chunk's gripper command history together with the current gripper state: we average the last $K{=}4$ gripper commands to obtain a continuous close/open intent score, combine command-based open intent with an aperture-based open score, and form the final payload flag as a clipped product of close intent and a soft closedness score derived from the measured gripper aperture.

\begin{figure*}[t]
\centering
% 1. HORIZONTAL SPACING
% 3pt gives nice breathing room between the "filmstrip" frames
\setlength{\tabcolsep}{3pt} 
\renewcommand{\arraystretch}{1.2}

\newcommand{\trajroot}{figures/extracted_frames}
% 2. IMAGE WIDTH
% Calculated to fit 10 images + label column within standard margins.
% If it's too wide, lower this to 0.075\linewidth
\newcommand{\imgw}{0.08\linewidth} 

% 3. TEXT WRAPPING COLUMN
% p{2.5cm} forces the text to wrap. \raggedright makes it look neat.
\begin{tabular}{p{2.5cm} @{\hspace{4pt}} *{10}{c}}
    \toprule
    \textbf{\textsc{Step} $T$} & 0 & 1 & 2 & 3 & 4 & 5 & 6 & 7 & 8 \\
    \midrule
    
    % Row 1
    \raggedright\scriptsize\textsc{External Camera} &
    \includegraphics[width=\imgw, valign=m]{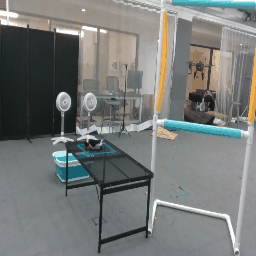} &
    \includegraphics[width=\imgw, valign=m]{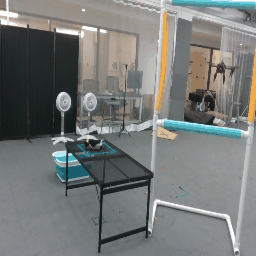} &
    \includegraphics[width=\imgw, valign=m]{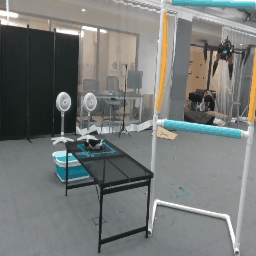} &
    \includegraphics[width=\imgw, valign=m]{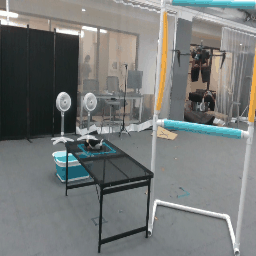} &
    \includegraphics[width=\imgw, valign=m]{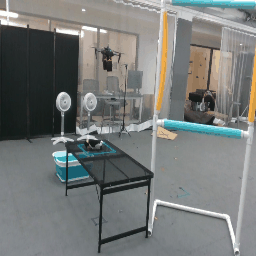} &
    \includegraphics[width=\imgw, valign=m]{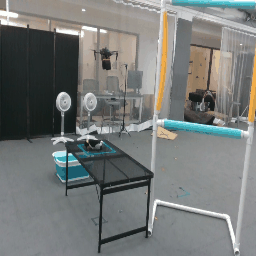} &
    \includegraphics[width=\imgw, valign=m]{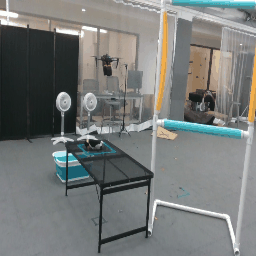} &
    \includegraphics[width=\imgw, valign=m]{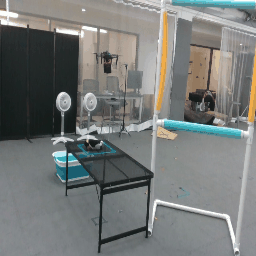} &
    \includegraphics[width=\imgw, valign=m]{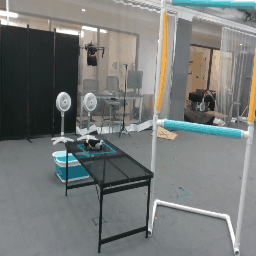} \\
    \addlinespace[4pt] % Vertical gap between rows

    % Row 2
    \raggedright\scriptsize\textsc{Forward Facing Camera} &
    \includegraphics[width=\imgw, valign=m]{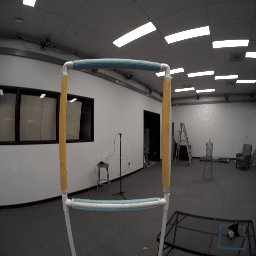} &
    \includegraphics[width=\imgw, valign=m]{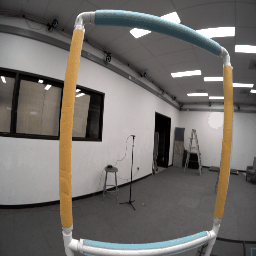} &
    \includegraphics[width=\imgw, valign=m]{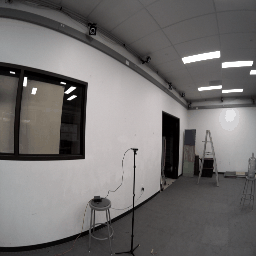} &
    \includegraphics[width=\imgw, valign=m]{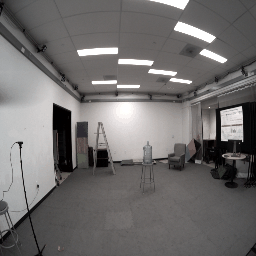} &
    \includegraphics[width=\imgw, valign=m]{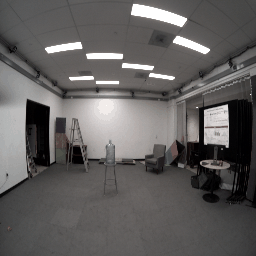} &
    \includegraphics[width=\imgw, valign=m]{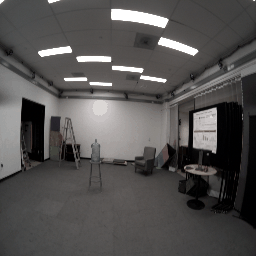} &
    \includegraphics[width=\imgw, valign=m]{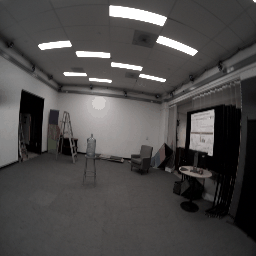} &
    \includegraphics[width=\imgw, valign=m]{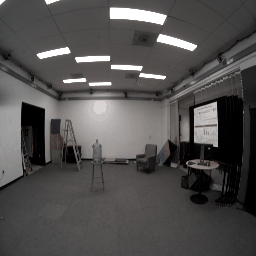} &
    \includegraphics[width=\imgw, valign=m]{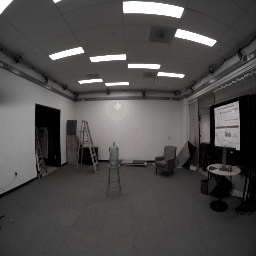} \\
    \addlinespace[4pt]

    % Row 3
    \raggedright\scriptsize\textsc{Downward Facing Camera} &
    \includegraphics[width=\imgw, valign=m]{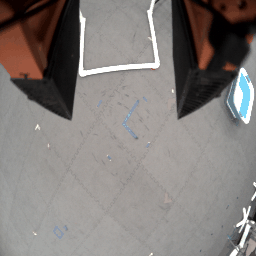} &
    \includegraphics[width=\imgw, valign=m]{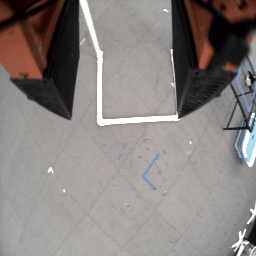} &
    \includegraphics[width=\imgw, valign=m]{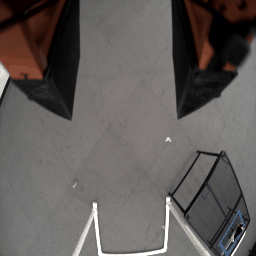} &
    \includegraphics[width=\imgw, valign=m]{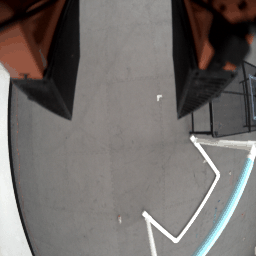} &
    \includegraphics[width=\imgw, valign=m]{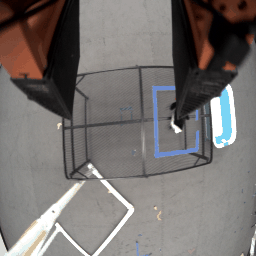} &
    \includegraphics[width=\imgw, valign=m]{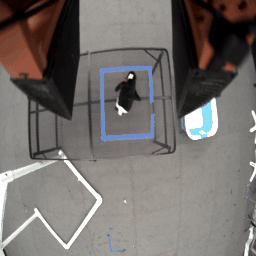} &
    \includegraphics[width=\imgw, valign=m]{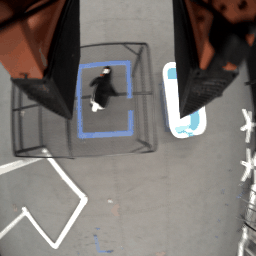} &
    \includegraphics[width=\imgw, valign=m]{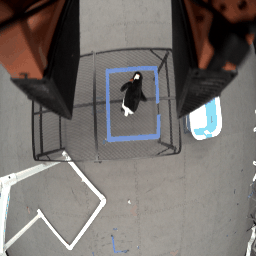} &
    \includegraphics[width=\imgw, valign=m]{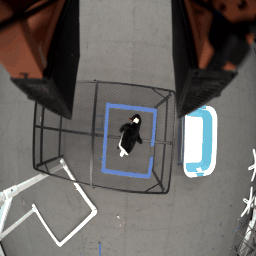} \\
    \addlinespace[4pt]

    % Row 4
    \raggedright\scriptsize\textsc{Synthetic Forward Downward Camera} &
    \includegraphics[width=\imgw, valign=m]{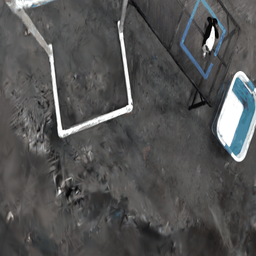} &
    \includegraphics[width=\imgw, valign=m]{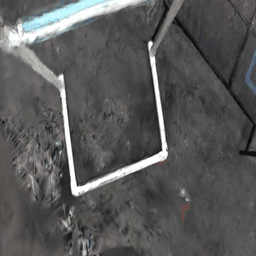} &
    \includegraphics[width=\imgw, valign=m]{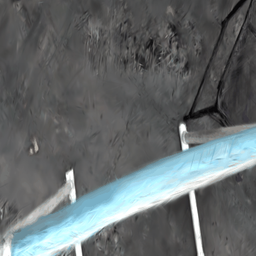} &
    \includegraphics[width=\imgw, valign=m]{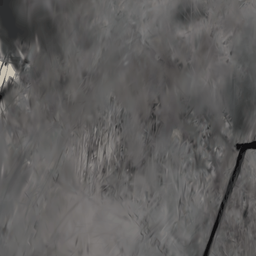} &
    \includegraphics[width=\imgw, valign=m]{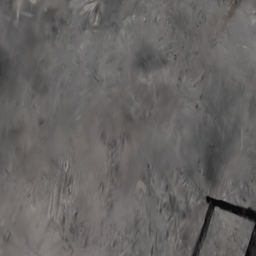} &
    \includegraphics[width=\imgw, valign=m]{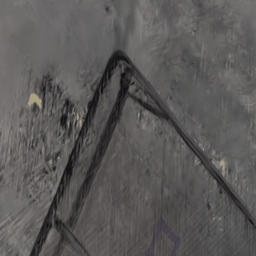} &
    \includegraphics[width=\imgw, valign=m]{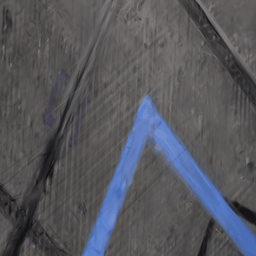} &
    \includegraphics[width=\imgw, valign=m]{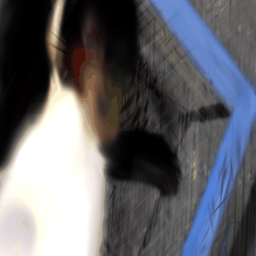} &
    \includegraphics[width=\imgw, valign=m]{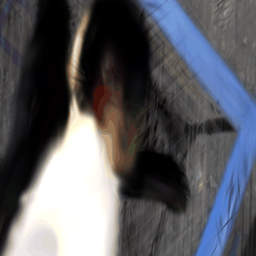} \\
    \addlinespace[4pt]

    % Row 5
    \raggedright\scriptsize\textsc{Synthetic Forward Facing Camera} &
    \includegraphics[width=\imgw, valign=m]{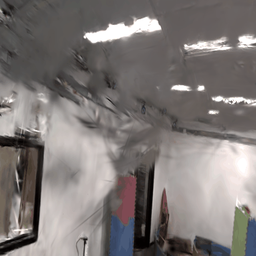} &
    \includegraphics[width=\imgw, valign=m]{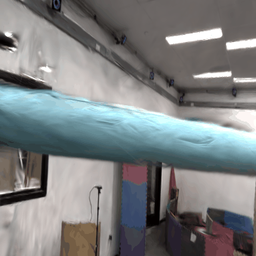} &
    \includegraphics[width=\imgw, valign=m]{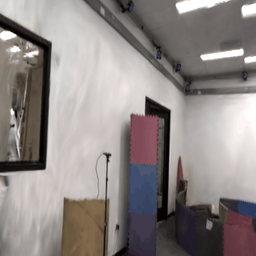} &
    \includegraphics[width=\imgw, valign=m]{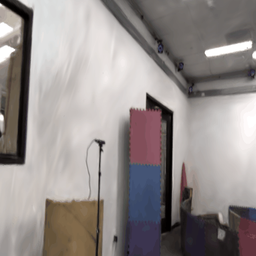} &
    \includegraphics[width=\imgw, valign=m]{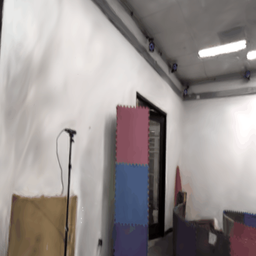} &
    \includegraphics[width=\imgw, valign=m]{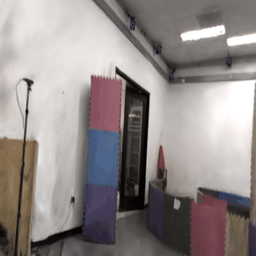} &
    \includegraphics[width=\imgw, valign=m]{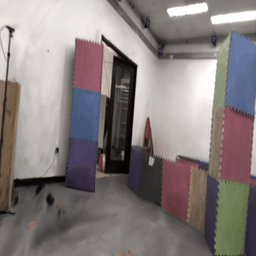} &
    \includegraphics[width=\imgw, valign=m]{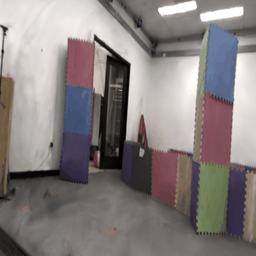} &
    \includegraphics[width=\imgw, valign=m]{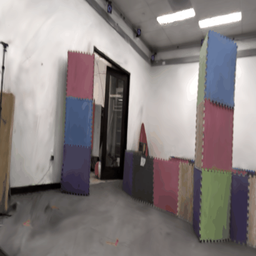} \\

    \bottomrule
\end{tabular}

\caption{\textbf{Trajectory visualization for the left gate task.}The top three rows display real-world RGB observations from external, forward, and downward cameras. The bottom two rows show corresponding synthetic views rendered from Gaussian Splats.}
\label{fig:trajectory_filmstrip}
\end{figure*}

\subsection{AirVLA Generalization}
A natural question after fine-tuning is how well the policy generalizes to out-of-distribution (OOD) settings. We probe this along three axes in our pick-and-place task: (i) novel object generalization, (ii) novel language prompt generalization, and (iii) novel environment generalization. We emphasize that these evaluations are not meant to demonstrate robust open-world aerial manipulation. Rather, they test whether a manipulation-pretrained VLA can retain useful behavior after transfer to an underactuated aerial platform.

% Generalization to novel objects
We ran experiments varying the object and fixing the language prompt in the pick-and-place task: ``pick up the stuffed animal and put it in the blue bucket''. When changing the stuffed animal from a penguin to a goomba (see Figure \ref{fig:ood}), we noticed that the policy would reliably attempt to pick the novel object as long as it was initially within the training region on the table.  This suggests that the fine-tuned policy transfers a grasping behavior beyond the exact training sequence, but that task performance is tightly linked to a workspace prior such as familiar grasping regions on the table. 

% Generalization to novel language tasks
To evaluate a wider range of novel objects, we varied the task prompt to match the object: ``pick up the \textless OBJECT \textgreater and put it in the blue bin.'' We detail the results of these evaluations in the main paper (see Table \ref{tab:ood_robustness_lean}). The objects used for these experiments are shown in Figure \ref{fig:ood}. An interesting qualitative observation from these evaluations is that the policy would attempt to pick objects in the training region even when no prompt was supplied. This indicates that part of the behavior may be driven by a strong visuomotor prior (i.e. ``if an object is present, attempt to grasp it'') rather than language grounding. 

% Generalization to novel environments
Naturally, such behavior begs the question of how well would the policy perform in the presence of distractor objects. To test this, we repeated the penguin grasp task with an additional out-of-distribution object within the training region. We noticed that regardless of the relative positions within the training region, the policy moved the drone to attempt grasps on the penguin rather than the distractor. This shows that fine-tuning can still induce useful object discrimination in manipulation, even when overall OOD robustness remains limited.

Overall, we view these results as evidence of partial transfer rather than broad generalization. The manipulation OOD results suggest that the pretrained VLA can still produce useful outputs under variation in object class, while the navigation OOD results in the main paper show substantial overfitting to the demonstration data and highlight the difficulty of cross-embodiment transfer when the pretrained model has not encountered a similar task. This interpretation is consistent with recent results showing that VLA models do not typically exhibit generalized zero-shot behavior after supervised fine-tuning, especially in the small-data regime \citep{fei2025libero,zhou2025libero}. Instead, a more realistic promise of VLAs is that supervised fine-tuning can enable few-shot adaptation to related new tasks.

We hypothesize that the limitations observed in these probes are partly driven by data scarcity (about 270 teleoperation demonstrations across all tasks), which can encourage reliance on workspace priors and habitual action routines. As larger and more diverse robot datasets become available, spanning broader embodiments, objects, and environments, we expect models such as AirVLA to improve generalization across each of the settings tested here. In future work, we will test this hypothesis by scaling demonstrations and measuring OOD performance as a function of data diversity and coverage.

\end{document}